\documentclass[letterpaper, 10 pt, conference]{ieeeconf}  

\IEEEoverridecommandlockouts                              

\usepackage{graphics} 
\usepackage{epsfig} 
\usepackage{mathptmx} 
\usepackage{times} 
\usepackage{amsmath} 
\usepackage{amssymb}  
\usepackage[T1]{fontenc}
\usepackage{multirow}

\title{\LARGE \bf
RS2AD: End-to-End Autonomous Driving Data Generation from Roadside Sensor Observations
}

\author{Ruidan Xing$^{1,2,3}$, Runyi Huang$^{1,2,4}$, Qing Xu$^{1,2}$ and Lei He$^{1,2,*}$ 
\thanks{*Corresponding author  {\tt\small helei2023@tsinghua.edu.cn}}
\thanks{$^{1}$School of Vehicle and Mobility, Tsinghua
University, Beijing, 100084, China.
       }%
\thanks{$^{2}$State Key Laboratory of Intelligent Green Vehicle and
Mobility, Tsinghua University, Beijing, 100084, China.
        }%
\thanks{$^{3}$School of Instrumentation and Optoelectronic Engineering, BeiHang University, Beijing, 100191, China.
        }%
\thanks{$^{4}$Department of Automation, Tsinghua University, Beijing, 100084, China.
       }%
\thanks{This study is supported by the National Key R\&D Program of China, Project "Development of Large Model Technology and Scenario Library Construction for Autonomous Driving Data Closed-Loop" (Grant No. 2024YFB2505501).}
}

\begin{document}

\maketitle
\thispagestyle{empty}
\pagestyle{empty}


\begin{abstract}

End-to-end autonomous driving solutions, which process multi-modal sensory data to directly generate refined control commands, have become a dominant paradigm in autonomous driving research. However, these approaches predominantly depend on single-vehicle data collection for model training and optimization, resulting in significant challenges such as high data acquisition and annotation costs, the scarcity of critical driving scenarios, and fragmented datasets that impede model generalization. To mitigate these limitations, we introduce RS2AD, a novel framework for reconstructing and synthesizing vehicle-mounted LiDAR data from roadside sensor observations. Specifically, our method transforms roadside LiDAR point clouds into the vehicle-mounted LiDAR coordinate system by leveraging the target vehicle’s relative pose. Subsequently, high-fidelity vehicle-mounted LiDAR data is synthesized through virtual LiDAR modeling, point cloud classification, and resampling techniques. To the best of our knowledge, this is the first approach to reconstruct vehicle-mounted LiDAR data from roadside sensor inputs.  
Extensive experimental evaluations demonstrate that incorporating the data generated by the RS2AD method (the RS2V-L dataset) into model training as a supplement to the KITTI dataset can significantly enhance the accuracy of 3D object detection and greatly improve the efficiency of end-to-end autonomous driving data generation.  These findings strongly validate the effectiveness of the proposed method and underscore its potential in reducing dependence on costly vehicle-mounted data collection while improving the robustness of autonomous driving models.

\end{abstract}

\begin{figure*}[thpb]
  \centering
  \includegraphics[scale=0.58]{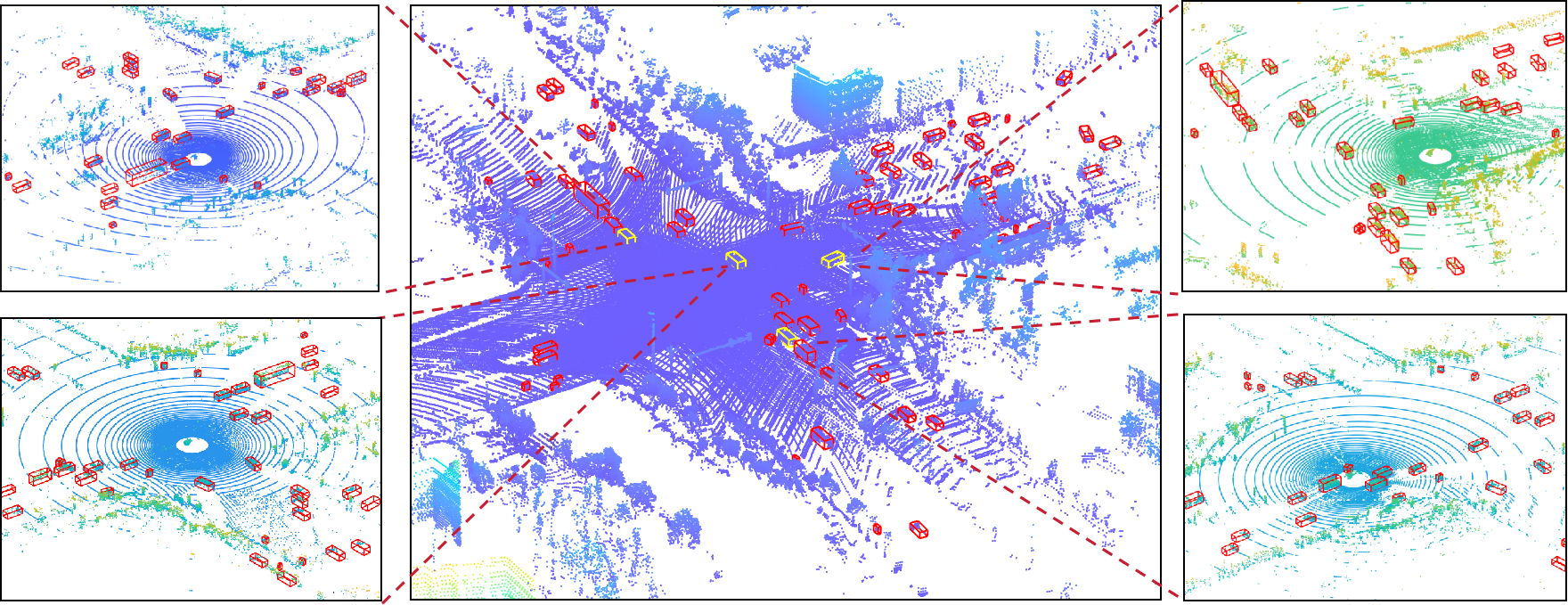}
  \caption{The proposed method generates vehicle-mounted LiDAR data for the four corresponding vehicles in the scene based on roadside LiDAR observations. The central section of the figure illustrates the data collected by the roadside LiDAR, while the left and right sections depict the vehicle-mounted LiDAR data for the vehicles highlighted within the yellow annotation boxes.}
  \label{Fig.1}
\end{figure*}

\section{INTRODUCTION}
End-to-end autonomous driving models, exemplified by Tesla FSD, aim to fully neuralize core autonomous driving algorithms\cite{a1}, significantly reducing the reliance on manually coded rule-based logic. These models have gradually become the mainstream trend in autonomous driving technology\cite{a7,a8}. However, current end-to-end autonomous driving systems still face several challenges, including high data collection and annotation costs, scarcity of high-value driving scenarios, and difficulties in effectively mining such scenarios\cite{a10}. Additionally, data silos arise as individual automakers restrict data loops to their own vehicle models, limiting cross-vehicle and cross-scenario adaptability and hindering model generalization.

To address these challenges, we propose a novel method to reconstruct vehicle-mounted LiDAR data from roadside sensor data, leveraging roadside sensors to generate high-precision dynamic and static 3D scene reconstructions. This enables the creation of sensor data adaptable to different vehicle models and sensor configurations across various automakers. Our method offers several key advantages: (1) All-weather, all-operation condition adaptation, ensuring stable performance across diverse environments and road conditions; (2) Support for arbitrary vehicle models and sensor configurations, enhancing model adaptability; (3) Low-cost, high-efficiency data generation, significantly reducing the cost of autonomous driving data collection and annotation; (4) Consistency between generated and real-world data, effectively mitigating the Sim2Real domain adaptation issue\cite{a12}, thereby improving training stability and generalization.

Specifically, our approach first models vehicle-mounted LiDAR sensors based on roadside LiDAR point clouds, incorporating the relative pose of the target vehicle within the scene. The roadside LiDAR data is then mapped to the vehicle-mounted LiDAR coordinate system. Subsequently, for each virtual LiDAR ray, a plane fitting approach is applied within its cone of vision using the roadside LiDAR point cloud. The intersection of the ray and the fitted plane is designated as the corresponding LiDAR point. If both ground points and non-ground points exist within the ray’s field of view, the non-ground points are prioritized for plane fitting to determine the final LiDAR sampling point, ensuring data accuracy and consistency.

Our Main Contributions:
\begin{itemize}
    \item The first method to reconstruct vehicle-mounted LiDAR sensor data from roadside sensor data.
    \item A novel LiDAR sampling and modeling approach that eliminates the Sim2Real adaptation issue, enhancing data realism and consistency.
    \item Comprehensive experiments validating the effectiveness of the proposed method, demonstrating its capability to efficiently generate end-to-end autonomous driving training data for different vehicle models.
\end{itemize}

\section{RELATED WORK}
The rapid advancement of autonomous driving research is fundamentally driven by the availability of large-scale, high-quality datasets. However, the high cost associated with collecting point cloud data, coupled with severe data silos, poses significant challenges for researchers in acquiring sufficient real-world data for training deep learning models. Consequently, generating reliable point cloud data for end-to-end autonomous driving training has emerged as a key research focus in recent years. Existing approaches for point cloud data generation can be broadly categorized into two main types: real-vehicle-based generation and environment simulation.

\subsection{Real-Vehicle-Based Generation}

Acquiring datasets from real vehicles is the most intuitive approach, and the release of various large-scale autonomous driving datasets has significantly propelled research in this field. For instance, SYNTHIA \cite{a13} and Cityscapes \cite{a14} primarily provide 2D annotations for images, whereas datasets such as KITTI \cite{a15}, nuScenes \cite{a16}, and Waymo \cite{a17} offer multi-modal data, including camera images and LiDAR point clouds, thereby serving as valuable resources for autonomous driving perception tasks. However, the collection and annotation processes are costly and labor-intensive, and these datasets often fail to comprehensively cover rare and complex long-tail scenarios. This limitation has motivated researchers to explore alternative data generation methodologies.

Early approaches for point cloud generation predominantly relied on Range Image Representation. However, such methods suffer from accuracy limitations in large-scale and complex environments. Caccia et al. \cite{a18} proposed a deep generative model-based method, where LiDAR scans were projected onto a 2D spherical point map, followed by the application of generative adversarial networks (GANs) and variational autoencoders (VAEs) for unconditional LiDAR point cloud generation. To enhance the quality of the generated 2D signals, absolute position information was incorporated. Building upon this foundation, Zyrianov et al. \cite{a19} introduced LiDARGen, which employs a point-based diffusion model for random denoising in an equirectangular projection view, thereby improving the realism and scalability of point cloud generation. However, the range image representation is inherently ego-centric, often failing to preserve the curved structures within LiDAR data and exhibiting limitations in recovering point clouds at certain viewing angles.

To address these shortcomings, UltraLiDAR \cite{a20} leverages a Bird's Eye View (BEV) voxel grid representation, enabling the encoding of geometric structures and occlusion relationships within point clouds to facilitate point cloud completion, generation, and manipulation. While UltraLiDAR excels in completing sparse point clouds, it lacks the flexibility needed for generating point clouds from arbitrary perspectives. LiDARDM \cite{a21} further advances the field by integrating implicit diffusion models with physical simulations to generate temporally consistent 4D LiDAR sequences, which are particularly useful for training and evaluating autonomous driving models. However, these methods predominantly rely on vehicle-mounted LiDAR data. Due to the height constraints of the vehicle itself and occlusions within dense traffic environments, such methods struggle to overcome limitations in field-of-view coverage.  

In contrast, data generation based on roadside sensors is inherently free from vehicle height constraints and occlusion issues, allowing for the acquisition of more comprehensive point cloud information \cite{a31}. Accordingly, our proposed method leverages real-world roadside LiDAR data, which not only preserves the curved structural characteristics of point cloud data but also mitigates mutual occlusion effects between vehicles, thereby providing a more complete and reliable dataset for autonomous driving applications.

\subsection{Environmental Simulation}

An alternative approach to point cloud generation involves synthesizing LiDAR data within simulation environments, such as CARLA \cite{a22} and Gazebo \cite{a23}, which utilize ray-casting techniques to simulate LiDAR sensing. These platforms emit rays from a virtual sensor origin and compute their intersections with geometric surfaces to construct point clouds. CARLA \cite{a22} and AirSim \cite{a24} provide high-fidelity environments populated with both static (e.g., buildings, trees) and dynamic (e.g., vehicles, pedestrians) objects, enabling the generation of structured point clouds via virtual LiDAR sensors. However, the reliance on predefined 3D assets inherently constrains scene diversity and limits the ability to accurately replicate complex real-world dynamics. Furthermore, the well-documented \textit{sim-to-real} gap poses a significant challenge, impeding the direct transferability of models trained on synthetic data to real-world scenarios.

To mitigate \textit{sim-to-real} discrepancies, various data-driven approaches have been proposed. For instance, LiDARSim \cite{a25} leverages deep learning techniques to correct simulation artifacts, while NeRF-based methods \cite{a26}\cite{a27} enhance the fidelity of asset reconstruction, refining the alignment between synthetic and real point cloud distributions. Despite these advancements, such methods remain dependent on simulation frameworks or reconstructed 3D assets, making it difficult to fully eliminate \textit{sim-to-real} inconsistencies. Additionally, many of these techniques require substantial amounts of real-scanned data for fine-tuning, leading to considerable data acquisition costs.

In contrast, this study departs from conventional simulation-based paradigms by leveraging real-world roadside sensor data to directly generate LiDAR point clouds. This approach not only reduces \textit{sim-to-real} discrepancies but also enhances model generalization in real-world environments, overcoming the limitations associated with synthetic data generation.

\section{METHOD}

\begin{figure*}[thpb]
  \centering
  \includegraphics[scale=0.45]{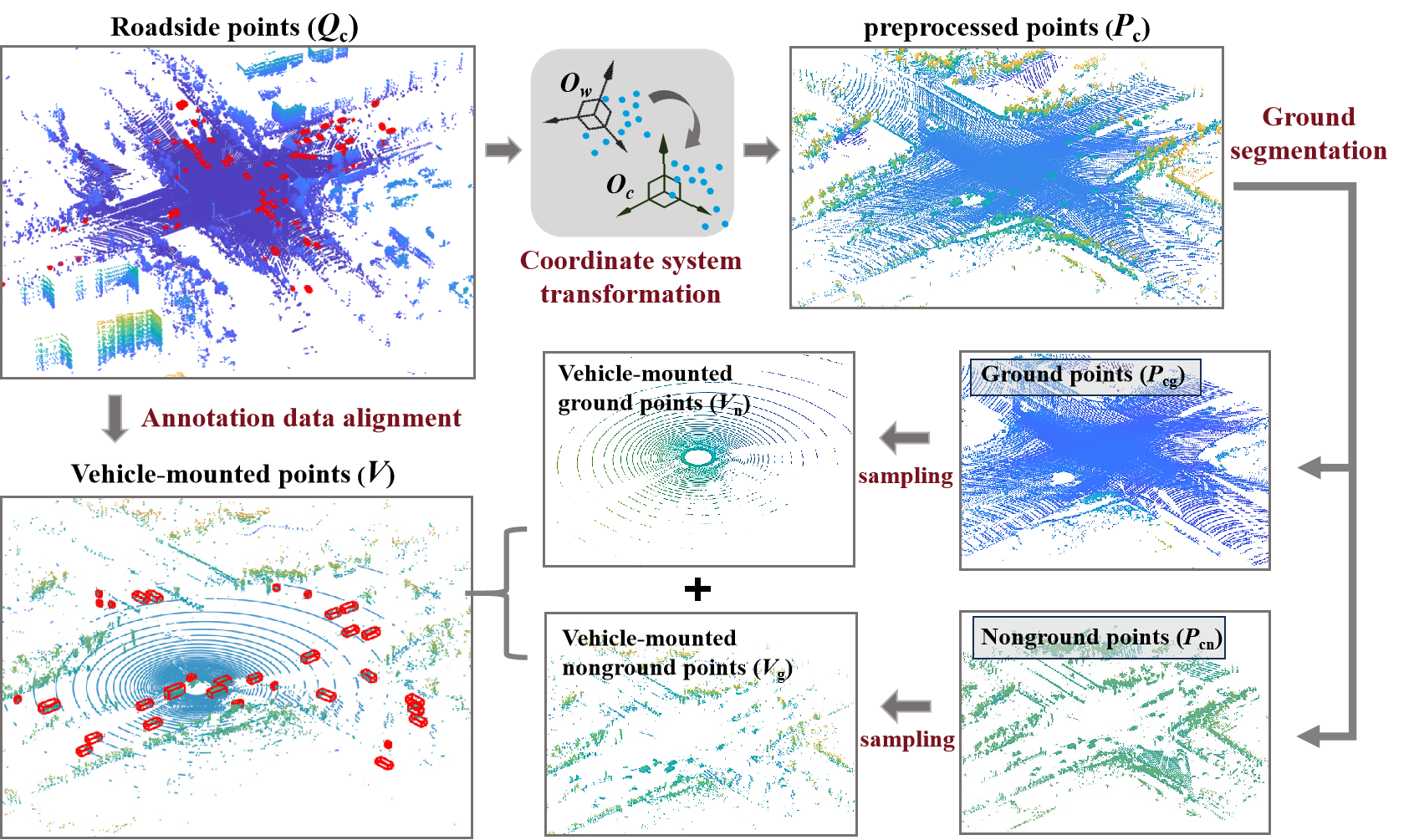}
  \caption{The overall technical architecture of RS2AD.}
  \label{Fig.2}
\end{figure*}

This section provides a detailed exposition of the technical implementation of the RS2AD system. Specifically, Section 3.1 presents the coordinate system transformation and data alignment process, which maps roadside LiDAR data to the vehicle-mounted coordinate frame based on real-time pose estimation. Section 3.2 introduces the methodology for generating virtual LiDAR data at the vehicle end. To achieve accurate scene reconstruction, we employ Patchwork++ \cite{a28} for semantic segmentation of object and ground point clouds. The segmented non-ground and ground point clouds are modeled separately, facilitating the synthesis of vehicle-mounted LiDAR point clouds through a virtual radar ray-tracing model. The overall architecture of the proposed method is illustrated in Fig. \ref{Fig.2}.

\subsection{Roadside-to-Vehicle Coordinate System Transformation and Data Alignment}

Let the original roadside LiDAR point cloud dataset be denoted as \(Q_w\), with all point coordinates expressed in the world coordinate system \(O_w\text{-}x_wy_wz_w\). This dataset is semantically annotated with scene objects, including their categories, three-dimensional geometric dimensions, centroid coordinates \(X_w\) in the world coordinate system, and corresponding rotation vectors \(\theta_w\).

To reconstruct vehicle-mounted LiDAR data, we first identify a target vehicle of interest from the annotated dataset. Let its pose parameters in the world coordinate system be given by \((X_{wc}, \theta_{wc})\). Utilizing the relative transformation \(\Delta T\) between the vehicle-mounted LiDAR and the target vehicle, we establish the vehicle-mounted LiDAR coordinate system \(O_c\text{-}x_cy_cz_c\). Following the standard coordinate system conventions of the KITTI dataset, the \(z\)-axis is defined as perpendicular to the ground and oriented upward, the \(x\)-axis points forward in the vehicle's direction of motion, and the \(y\)-axis extends laterally to the left, aligning with the vehicle's coordinate frame.

Based on this transformation framework, we derive a rigid-body transformation model that maps the roadside LiDAR data from the world coordinate system to the vehicle-mounted LiDAR coordinate system. The mathematical formulation of this transformation is given by:
\begin{align}
R&=\text{Rodrigues}(\theta_{wc})^{-1}\tag{1}\\
T&=-RX_{wc}-\Delta T\tag{2}
\end{align}
where \(\text{Rodrigues}(\cdot)\) represents the Rodrigues rotation formula.

Building upon the rigid-body transformation model, precise alignment of the point cloud data from the global coordinate system to the vehicle-mounted radar coordinate system is achieved, yielding the transformed point cloud \(Q_c\) in the vehicle-mounted radar frame. Simultaneously, corresponding transformations are applied to the annotation information to ensure consistency across coordinate systems. Throughout this process, the object's category and geometric dimensions remain unchanged, while its pose \((X_c, \theta_c)\) in the vehicle-mounted radar coordinate system is computed using the following expressions:
\begin{align}
X_c&=RX_w + T\tag{3}\\
\theta_c&=\text{InvRodrigues}(R\cdot\text{Rodrigues}(\theta_w))\tag{4}
\end{align}

\subsection{Vehicle-Mounted LiDAR Data Generation Method}

To facilitate accurate point cloud synthesis, we first establish a virtual LiDAR model. Based on the LiDAR parameters used in the KITTI dataset, we construct a vehicle-mounted virtual LiDAR system within a spherical coordinate framework. The system operates within a sensing range of 0.5 m to 100 m, covering a horizontal field of view of \([0, 360^{\circ}]\) and a vertical field of view spanning \([88^{\circ}, 114^{\circ}]\). The horizontal and vertical resolutions are defined by \(m\) and \(k\) discrete lines, respectively, resulting in a total of \(n = m\times k\) rays per scan cycle.

The virtual LiDAR emits a set of rays, denoted as \(L\), originating from the center of the spherical coordinate system and propagating in all directions. Each individual ray is indexed as \(l_{ij}\), where \(i = 0, 1, 2,\ldots, m-1\) represents the horizontal division index, and \(j = 0, 1, 2,\ldots, k-1\) represents the vertical division index. The horizontal azimuth angle \(\varphi_{ij}\) and vertical elevation angle \(\theta_{j}\) for the \((i, j)\)-th ray are computed as follows:
\[
\begin{cases}
\varphi_{ij}=\frac{i}{m}\times360^{\circ}\\
\theta_{j}=88^{\circ}+\frac{j}{k}\times26^{\circ}
\end{cases}\tag{5}
\]

\begin{figure}[thpb]
  \centering
  \includegraphics[scale=0.6]{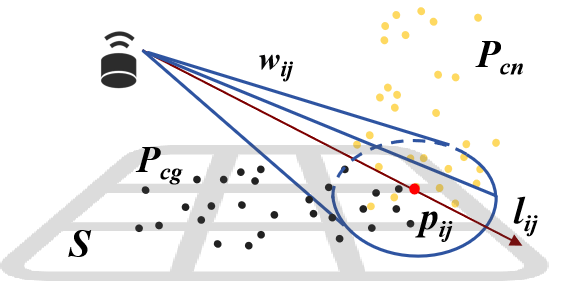}
  \caption{Schematic Representation of LiDAR Ray \(l_{ij}\) and Corresponding View Frustum \(w_{ij}\).}
  \label{Fig.3}
\end{figure}

Subsequently, the complete ray set \(L\) is defined as \(L = \{l_{ij}(\varphi_{ij}, \theta_{j}): i = 0, 1, 2,\ldots, m-1; j = 0, 1, 2,\ldots, k-1\}\). Taking the origin of the spherical coordinate system as the vertex and each ray in \(L\) as its central axis, a corresponding cone-shaped view frustum is constructed. The angular resolution of the LiDAR in the vertical direction, given by \((k / 26)^{\circ}\), determines the cone angle of each frustum. This results in a collection of view frustums denoted as \(W = \{w_{ij}: i = 0, 1, 2,\ldots, m-1; j = 0, 1, 2,\ldots, k-1\}\), where each individual frustum \(w_{ij}\) is associated with its corresponding ray \(l_{ij}\), as illustrated in Fig. \ref{Fig.3}.

Upon the completion of the vehicle-mounted virtual LiDAR model construction, pre-processing and ground segmentation operations are applied to the input point cloud data \(Q_c\). Let the initial point cloud be represented as \(Q_c=\{q_1, q_2,\ldots, q_N\}\), which is subsequently transformed into its spherical coordinate representation. Based on the predefined sensing range of the virtual LiDAR model, points falling outside this range are discarded, yielding the pre-processed point set \(P_c = \{p_1, p_2,\ldots, p_N\}\), as depicted in the upper-right corner of Fig. \ref{Fig.2}. Each point in this set is expressed as \(p_s=(\rho_s, \varphi_s, \theta_s)\).

Next, Patchwork++ \cite{a28} is employed to perform ground segmentation on \(P_c\), partitioning it into a non-ground point cloud \(P_{cn}\) and a ground point cloud \(P_{cg}\), as illustrated in the lower-right corner of Fig. \ref{Fig.2}. For any point \(p_i\) in the non-ground point cloud \(P_{cn}\), its corresponding view frustum is determined based on its polar angle \(\theta_i\) and azimuth angle \(\varphi_i\). Within each frustum, plane fitting is performed separately for the contained point cloud. In contrast, the ground point cloud \(P_{cg}\) undergoes a global plane fitting process via the least-squares method, as shown in Fig. \ref{Fig.3}.

To associate each non-ground point \(p_i \in P_{cn}\) with its corresponding ray and view frustum, its indices \((i, j)\) are determined based on its polar and horizontal angular divisions as follows:
\[
\begin{cases}
i = \text{round}(\frac{m}{360^{\circ}}\varphi_i) \bmod m\\
j = \text{round}(\frac{k}{26^{\circ}}\theta_i) \bmod k
\end{cases}\tag{6}
\]

Denote the point set corresponding to the ray \(l_{ij}\) and its associated view frustum \(w_{ij}\) as \(P_{ij}\), where  
\(
P_{ij}=\{p_s\in P\mid i = \text{round}(m\varphi_i/360^{\circ})\bmod m, \quad j=\text{round}(k\theta_i/26^{\circ})\bmod k\}.
\) 
In scenarios where the roadside point cloud is relatively sparse, to ensure sufficient density in the generated non-ground point cloud, the angular range of the view frustum can be expanded to approximately 2–3 times its original value, thereby increasing the number of points encompassed within each frustum. For each non-empty point set \(P_{ij}\), a plane \(S_{ij}\) is fitted using the least-squares method. Subsequently, based on the ray-tracing model, the intersection point \(p_{ij}\) between the corresponding ray \(l_{ij}\) and the fitted plane \(S_{ij}\) is computed. The resulting intersection point \(p_{ij}\) serves as the sampling point of the vehicle-mounted virtual LiDAR along the direction \((\varphi_{ij}, \theta_{ij})\). All valid intersection points are aggregated into a set \(V_n\), which constitutes the vehicle-mounted non-road surface point cloud data.

For the ground point cloud \(P_{cg}\), a global plane fitting approach is employed to derive the fitted ground plane \(S\). To refine the selection of valid sampling points, rays \(l_{ij}\) that contain non-ground points within their respective view frustums but lack ground points are filtered out. For the remaining rays, their intersection points \(p_{ij}\) with the plane \(S\) are computed. All valid intersection points are then collected to form the set \(V_g\), representing the vehicle-mounted road surface point cloud data. Finally, merging \(V_n\) and \(V_g\) results in the complete vehicle-mounted LiDAR point cloud dataset \(V\), as illustrated in the lower-left corner of Fig. \ref{Fig.2}.

\begin{figure*}[thpb]
  \centering
  \includegraphics[scale=0.45]{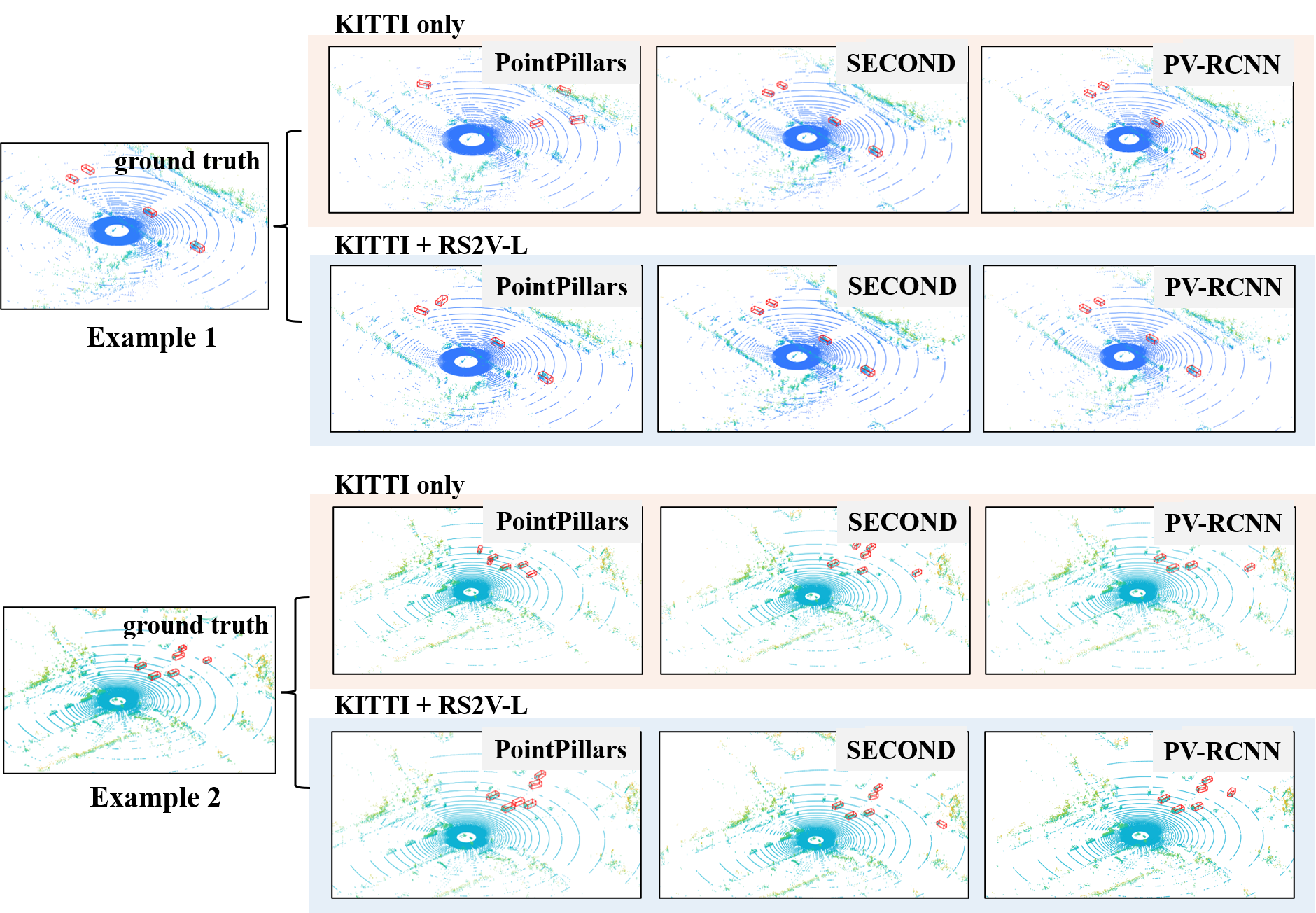}
  \caption{
Two scene examples are depicted, with the ground - truth illustrated on the far left. For each example, the first row presents the test results of distinct models (PointPillars, SECOND, PV-RCNN) trained solely on the KITTI dataset, while the SECOND row showcases the test outcomes of these models when trained on a combination of the KITTI dataset and the RS2V-L dataset. Through comparison between the two rows, a notable enhancement in detection accuracy is observable for the models trained with the mixed - dataset strategy.
}
  \label{Fig.4}
\end{figure*}

\section{EXPERIMENT}  

In this study, experiments are conducted using a roadside multi-sensor dataset independently collected by our team. This dataset is acquired through continuous data collection by four 128-line LiDAR sensors deployed at intersections, encompassing a total of 531 frames captured across 50 distinct time periods, including morning and evening rush hours, off-peak hours, and nighttime. Leveraging this dataset, we employ the RS2AD system to generate 6000 frames of vehicle-mounted point cloud data, covering more than 1000 unique vehicles. On average, each time period captures point cloud data for over 20 different vehicles. The RS2AD system not only enhances dataset acquisition efficiency by more than an order of magnitude but also significantly enriches data diversity. The generation results are illustrated in Fig. \ref{Fig.1}.  

To assess the effectiveness of the data generated by the RS2AD method in improving model training, we randomly select 4000 frames for experimentation. The dataset is partitioned into a training set comprising 3000 frames and a test set consisting of 1000 frames. A 3D object detection benchmarking platform is constructed based on OpenPCDet \cite{a29}, with the KITTI dataset \cite{a15} serving as the baseline for comparison. A comprehensive evaluation is conducted from two perspectives: (1) the overall performance enhancement of the model and (2) the specific improvements in complex traffic scenarios, such as densely populated intersections with high traffic flow.  

\subsection{Overall Model Performance Improvement}  



To evaluate the impact of the generated data on model performance, a comparative experiment was designed for models, including PointPillars\cite{a30}, SECOND, and PV-RCNN. Each model was trained under three configurations: solely on the KITTI dataset (KITTI: 3712), solely on the RS2V-L dataset, or on a mix of KITTI (3712 frames) and RS2V-L (3000 frames). Vehicle detection performance was evaluated on the KITTI validation set. The core evaluation metrics comprised average precision (AP) and AP\_R40 for Bird’s-Eye View (BEV) bounding boxes and 3D bounding boxes under Intersection over Union (IoU) thresholds of 0.7 and 0.5.

As summarized in TABLE \ref{tab:test_result_1}, for all models, the mix-training group (KITTI + RS2V-L) exhibited significant performance improvements. For instance, when compared to training with the KITTI dataset alone, the 3D bounding-box AP of the SECOND model under an IoU of 0.7 increased from 65.08 to 67.29 in the mix group. Compared to training with the RS2V-L dataset alone, the 3D bounding-box AP of PointPillars under an IoU of 0.5 rose from 58.61 to 84.32 in the mix group. These improvements in the AP and AP\_R40 metrics illustrate that mixed training with RS2V-L enhances model performance and generalization. Thus, the RS2AD framework serves as an effective data-augmentation and model-optimization tool, enhancing the robustness and reliability of autonomous-driving perception models.

\begin{table*}
\centering
\caption{Evaluation of KITTI Dataset Performance Under Different Training Data Configurations}
\label{tab:test_result_1}
\begin{tabular}{lcccccccccc}
\hline
\multirow{2}{*}{Model} & \multirow{2}{*}{Dataset} & \multicolumn{4}{c}{IoU: 0.7} & \multicolumn{4}{c}{IoU: 0.5} \\
\cline{3 - 10}
& & \multicolumn{2}{c}{AP} & \multicolumn{2}{c}{AP\_R40} & \multicolumn{2}{c}{AP} & \multicolumn{2}{c}{AP\_R40} \\
\cline{3 - 10}
& & bev & 3d & bev & 3d & bev & 3d & bev & 3d \\
\hline
\multirow{3}{*}{PointPillars} 
& Kitti & \textbf{79.27} & \textbf{65.87} & \textbf{79.83} & \textbf{67.44} & 84.65 & 84.29 & 87.93 & 86.48 \\
& RS2V-L & 54.21 & 42.22 & 54.33 & 41.16 & 59.72 & 58.61 & 60.27 & 59.27 \\
& Mix & 79.19 & 65.58 & 79.70 & 66.90 & \textbf{84.80} & \textbf{84.32} & \textbf{88.10} & \textbf{86.48} \\
\hline
\multirow{3}{*}{SECOND} 
& Kitti & \textbf{77.74} & 65.08 & 78.80 & 66.25 & 83.42 & 82.88 & 86.28 & 84.73 \\
& RS2V-L & 59.34 & 52.59 & 60.50 & 52.94 & 63.64 & 63.19 & 64.94 & 64.34 \\
& Mix & 77.43 & \textbf{67.29} & \textbf{78.95} & \textbf{67.62} & \textbf{83.59} & \textbf{83.21} & \textbf{86.50} & \textbf{84.97} \\
\hline
\multirow{3}{*}{PV-RCNN} 
& Kitti & \textbf{79.52} & 70.92 & \textbf{80.31} & 71.26 & \textbf{83.7} & \textbf{83.60} & \textbf{87.24} & \textbf{85.77} \\
& RS2V-L & 27.59 & 3.88 & 26.60 & 3.19 & 55.00 & 50.14 & 52.56 & 47.62 \\
& Mix & 77.66 & \textbf{71.50} & 78.78 & \textbf{71.36} & 81.90 & 81.81 & 85.19 & 84.65 \\
\hline
\end{tabular}
\end{table*}

\subsection{Optimization for Complex Scenarios}  
To evaluate the effectiveness of the RS2AD-generated data in enhancing model performance under complex scenarios and its applicability for data collection in extreme conditions, this study conducts experiments with models such as PointPillars, SECOND, and PV-RCNN. Each model is trained under three configurations: solely on the KITTI dataset, solely on the RS2V-L dataset, or on a mix of KITTI and RS2V-L. The evaluation focuses on model performance when tested on the RS2V-L test set.

As shown in TABLE \ref{tab:test_result_2}, for all models, mixed training (KITTI + RS2V-L) yields significant performance improvements. Take PointPillars as an example: compared to KITTI only training, its 3D detection-box average precision (AP) on the RS2V-L test set improves remarkably; compared to RS2V-L only training, the mix-training group also demonstrates notable enhancements. These AP metric improvements for 3D and Bird’s-Eye-View (BEV) detection boxes illustrate that incorporating RS2V-L for mixed training effectively strengthens the model’s capability to handle complex scenarios, thereby enhancing object-detection performance. Fig.\ref{Fig.4} displays the experiment’s visualization results, where the model trained on the mixed dataset shows a significantly improved object-detection ability, further confirming the positive impact of RS2AD-generated data.

\begin{table*}
\centering
\caption{Evaluation of RS2V-L Dataset Performance Under Different Training Data Configurations}
\label{tab:test_result_2}
\begin{tabular}{lcccccccccc}
\hline
\multirow{2}{*}{Model} & \multirow{2}{*}{Dataset} & \multicolumn{4}{c}{IoU: 0.7} & \multicolumn{4}{c}{IoU: 0.5} \\
\cline{3 - 10}
& & \multicolumn{2}{c}{AP} & \multicolumn{2}{c}{AP\_R40} & \multicolumn{2}{c}{AP} & \multicolumn{2}{c}{AP\_R40} \\
\cline{3 - 10}
& & bev & 3d & bev & 3d & bev & 3d & bev & 3d \\
\hline
\multirow{3}{*}{PointPillars} 
& Kitti & 14.31 & 9.31 & 8.76 & 0.81 & 34.09 & 28.02 & 31.16 & 24.54 \\
& RS2V-L & 77.09 & \textbf{61.58} & \textbf{77.47} & 60.55 & \textbf{88.44} & 86.19 & 89.70 & \textbf{88.68} \\
& Mix & \textbf{77.23} & 60.94 & 77.17 & \textbf{66.45} & 87.94 & 86.23 & \textbf{90.13} & 88.08 \\
\hline
\multirow{3}{*}{SECOND} 
& Kitti & 13.08 & 4.65 & 6.49 & 0.97 & 28.66 & 23.87 & 24.98 & 19.85 \\
& RS2V-L & 77.82 & \textbf{64.24} & \textbf{79.39} & \textbf{63.10} & \textbf{88.41} & \textbf{87.52} & \textbf{91.13} & \textbf{88.70} \\
& Mix & \textbf{78.01} & 61.21 & 79.36 & 60.58 & 88.29 & 86.69 & 90.56 & 88.47 \\
\hline
\multirow{3}{*}{PV-RCNN} 
& Kitti & 17.17 & 9.56 & 11.88 & 2.69 & 32.01 & 27.58 & 28.89 & 24.12 \\
& RS2V-L & 73.57 & 60.51 & 75.57 & 60.01 & 85.07 & 83.52 & 87.14 & 85.07 \\
& Mix & \textbf{78.22} & \textbf{66.15} & \textbf{79.70} & \textbf{65.07} & \textbf{88.01} & \textbf{86.68} & \textbf{89.28} & \textbf{88.73} \\
\hline
\end{tabular}
\end{table*}

The RS2AD framework proposed in this study leverages 531 frames of roadside LiDAR data to generate 6000 frames of vehicle-mounted LiDAR data, achieving a more than tenfold increase in data collection efficiency. Furthermore, the effectiveness of the RS2AD-generated data in improving model performance is systematically validated through two sets of comparative experiments.

In the overall performance evaluation, models such as PointPillars, SECOND, and PV-RCNN trained on the augmented dataset (KITTI + RS2V-L) overall outperform their counterparts trained solely on the KITTI dataset, across key metrics for both Bird’s-Eye-View (BEV) and 3D object detection under various Intersection over Union (IoU) thresholds. Taking SECOND as an illustrative example, under an IoU threshold of 0.7, the 3D average precision (AP) improves from 65.08 (KITTI-only) to 67.29 (Mix). At a more relaxed threshold of IoU = 0.5, the 3D AP rises from 83.42 to 83.59.

In the evaluation under complex scenarios—particularly at intersection-like environments assessed on the RS2V-L test set—the benefits of the augmented dataset are even more pronounced. Specifically, for PV-RCNN under IoU = 0.7, the BEV AP improves dramatically from 17.17 (KITTI only) to 78.22 (Mix), and the 3D AP increases from 9.56 to 66.15. Under IoU = 0.5, the BEV AP rises from 32.01 to 88.01, while the 3D AP climbs from 27.58 to 86.68.

In summary, the  RS2AD-generated data not only enhances the overall model performance but also substantially improves its effectiveness in complex and highly dynamic scenarios. This approach offers significant value for augmenting vehicle-mounted LiDAR datasets and facilitating data collection in extreme driving conditions.

\section{CONCLUSIONS}
This paper presents the RS2AD method, which effectively reconstructs and synthesizes vehicle-mounted LiDAR data from roadside sensor observations, offering an efficient and cost-effective data generation paradigm for autonomous driving applications. Experimental evaluations demonstrate that the data generated by RS2AD not only significantly enhances the overall performance of autonomous driving models but also exhibits remarkable adaptability in complex environments, such as intersections.

In contrast to conventional vehicle-mounted data acquisition approaches, RS2AD capitalizes on the "god’s eye view" advantage of roadside sensors to mitigate vehicle-to-vehicle occlusion while addressing domain adaptation challenges between simulated and real-world data. Furthermore, this method accommodates arbitrary vehicle models and sensor configurations, ensuring broad applicability across diverse autonomous driving scenarios.

Future research will focus on further optimizing the RS2AD framework, expanding its applicability to a wider range of scenarios and sensor modalities, and integrating multimodal data generation techniques. These advancements aim to provide more comprehensive support for the continued development of autonomous driving technology.

\addtolength{\textheight}{-12cm}   









\end{document}